\documentclass[letterpaper, 10 pt, conference]{ieeeconf}  

\IEEEoverridecommandlockouts                              

\title{\LARGE \bf
Dynamics Randomization Revisited: \\ A Case Study for Quadrupedal Locomotion   
}

\author{%
Zhaoming Xie$^{*1,2}$, %
Xingye Da$^{1}$, %
Michiel van de Panne$^{2}$, %
Buck Babich$^{1}$, %
Animesh Garg$^{1,3}$ 
\thanks{*Work done during an internship at NVIDIA}
\thanks{$^{1}$NVIDIA, $^{2}$University of British Columbia, $^3$University of Toronto, Vector Institute
        }%
}

\usepackage{amsfonts}
\usepackage{amsmath}
\usepackage{graphicx}
\usepackage{xcolor}
\usepackage{cite}
\usepackage{caption}
\usepackage{subcaption}
\usepackage[pagebackref=false,breaklinks=true,colorlinks,urlcolor=blue,citecolor=blue,linkcolor=blue,bookmarks=false]{hyperref}

\DeclareCaptionFormat{subfig}{\figurename~#1#2#3}
\DeclareCaptionSubType*{figure}

\newcommand{\norm}[1]{\left\lVert#1\right\rVert}

\begin{document}

\maketitle
\thispagestyle{empty}
\pagestyle{empty}

\begin{abstract}

Understanding the gap between simulation and reality is critical 
for reinforcement learning with legged robots, which are largely trained in simulation. 
However, recent work has resulted in sometimes conflicting conclusions 
with regard to which factors are important for success, including the role
of dynamics randomization.  
In this paper, we aim to provide clarity and understanding on
the role of dynamics randomization in learning robust locomotion policies for the Laikago quadruped robot. 
Surprisingly, in contrast to prior work with the same robot model, we find that direct sim-to-real transfer 
is possible without dynamics randomization or on-robot adaptation schemes.
We conduct extensive ablation studies in a sim-to-sim setting to understand the key issues underlying
successful policy transfer, including other design decisions that can impact policy robustness.
We further ground our conclusions via sim-to-real experiments with various gaits, speeds, and stepping frequencies. Additional Details: \href{https://www.pair.toronto.edu/understanding-dr/}{\texttt{pair.toronto.edu/understanding-dr/}}

\end{abstract}

\section{Introduction}

Deep reinforcement learning (RL) is increasingly successful in adoption as a feasible approach for synthesizing locomotion policies for legged robots. 
However, direct training on hardware is often impractical due to the sample efficiency of RL algorithms and unsafe exploration behaviors during the training phase.
Instead, a physics-based simulator is commonly employed during training.  
Moreover, the discrepancy between the simulator and the real world, also known as the ``reality gap,'' 
can cause direct sim-to-real transfer to fail. 
One way to combat this problem is to employ {\em dynamics randomization}, 
where parameters of the simulation system are randomized during training, in order to obtain policies that are robust to modeling errors. 
This has been used extensively in recent work in sim-to-real transfer for learned 
legged robot policies~\cite{2018-rss-sim2realquadruped, 2020-RSS-laikago, 2019-iros-sim2realbiped, 2019-science-sim2realAnyaml, 2020-science-blindQuadruped}.
However, conflicting observations have been made in other work where no dynamics randomization was needed for sim-to-real transfer~\cite{2019-CORL-cassie,2020-sim2real-noDyRand,2020-CoRL-ContactAdapt}. 

In this paper, we revisit dynamics randomization in detail, with the aim of providing an improved understanding
of when it should be used, grounded in sim-to-sim and sim-to-real experiments using the Unitree Laikago quadruped.  More specifically, we make the following contributions:
\begin{enumerate}
  \item We demonstrate that dynamics randomization is \textit{not necessary} for successful sim-to-real transfer in our settings, across multiple gaits and speeds, while robust to common types of perturbations. Note that the same robot model has been demonstrated to fail the direct sim-to-real test 
  for the same class of motions \cite{2020-RSS-laikago}.
  \item We identify and analyze particular design choices for which dynamics randomization is \textit{not sufficient} to enable sim-to-real success. 
  \item We evaluate the consequences of dynamics randomization in our setting. 
  Specifically, we find that unnecessary randomization of parameters can produce conservative policies
  with limited actual gains in robustness, 
  while randomization in truly problematic parameters bridges the reality gap.
\end{enumerate}

   \begin{figure}[t]
      \centering
      \includegraphics[width=0.90\columnwidth]{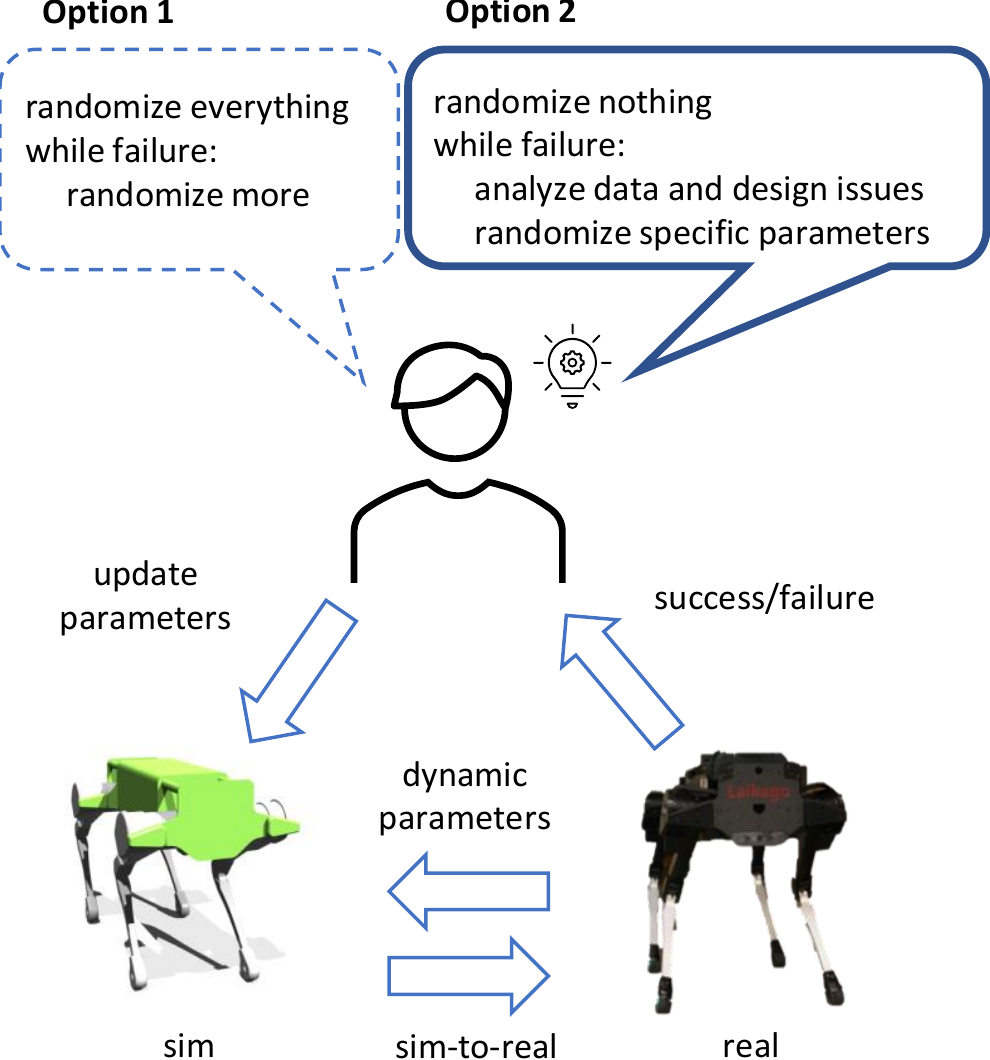}
      \caption{
      Dynamics randomization is often applied in an ad hoc fashion. We advocate for identifying and randomizing 
      parameters that matter and the importance of identifying other problematic control policy design issues.
      }
      \label{fig:teaser}
      \vspace{-15pt}
   \end{figure}
   
Empirical evidence in this paper illustrates that Domain Randomization is \textit{neither necessary nor sufficient} in some settings. We instead advocate for conservative application of dynamics randomization, i.e., such computationally intensive techniques should be used when there is a clearly identified need in sim-to-sim tests. Further, randomization should only be on parameters that matter, which requires domain insight. 
Fig.~\ref{fig:teaser} shows the naive approach and the advocated approach.
\section{Related Work}

\subsection{Quadrupedal Locomotion}
Model-based approaches such as trajectory optimization and model predictive control are commonly used to achieve agile locomotion skills on quadrupeds, typically making use of a low-dimensional model, e.g., \cite{2018-iros-cheetah3, 2013-icra-starlETH, 2020-iros-HyQ}. Deep RL has also been used to train quadrupedal controllers in simulation and transfer to a physical robot, e.g., \cite{2018-rss-sim2realquadruped, 2019-science-sim2realAnyaml,2020-icra-constraintQuadruped, 2020-RSS-laikago}.

\subsection{Sim-to-real for Robot Control}
The reality gap often prevents the success of direct sim-to-real transfer. Dynamics randomization describes the randomization of parameters such as masses and inertial moments of robot links, as well as parameters that govern control latency, actuator response, etc.  Such parameters are randomized during training in order to obtain policies that are robust to modeling errors, as first proposed to solve robotic arm pushing tasks \cite{2018-icra-dynamicRand} and later used for various sim-to-real studies involving manipulation \cite{2019-arxiv-shadowHand} and locomotion \cite{2018-rss-sim2realquadruped, 2019-science-sim2realAnyaml, 2020-rss-RNNCassie}. This technique complements the collection of on-robot data for online adaptation~\cite{2020-RSS-laikago, 2019-iros-sim2realbiped} or system identification in the form of improved simulator accuracy~\cite{2019-arxiv-BayesSim, 2020-arxiv-groundedSim}. At the same time, multiple results also demonstrate successful sim-to-real transfer without dynamics randomization via appropriate design choices, e.g.,~\cite{thananjeyan2017multilateral, 2019-CORL-cassie, 2020-arxiv-noDyRand, 2020-sim2real-noDyRand, 2020-CoRL-ContactAdapt}. 

\subsection{Robust Control}
Control policies obtained with a single model are often susceptible to modeling errors or noise, even for a simple linear quadratic regulator \cite{2012-acc-robust_policy}. Robust policies can be obtained via model ensembling, where a distribution of models is used for control synthesis, e.g.,\cite{2007-iros-CASTRO, 2015-iros-ensembleCIO}. Perturbations can also be introduced during optimization to obtain robust behaviors, e.g.,\cite{2012-CDC-leggedUncertainties, 2019-icra-highLevelRL, 2017-iros-adversarialRL}. Further two stage models achieve robustness through reference tracking of a state trajectory obtained from an open-loop policy rollout with feedback control~\cite{harrison2017adapt}.

In this work, we employ the Laikago robot, which has been used for prior sim-to-real work~\cite{2020-RSS-laikago,2020-CoRL-ContactAdapt}, and we explore in detail the role of dynamics randomization and the importance of appropriate design choices.
\section{Control Policies}

\begin{figure}[t]
  \centering
  \includegraphics[width=0.95\columnwidth]{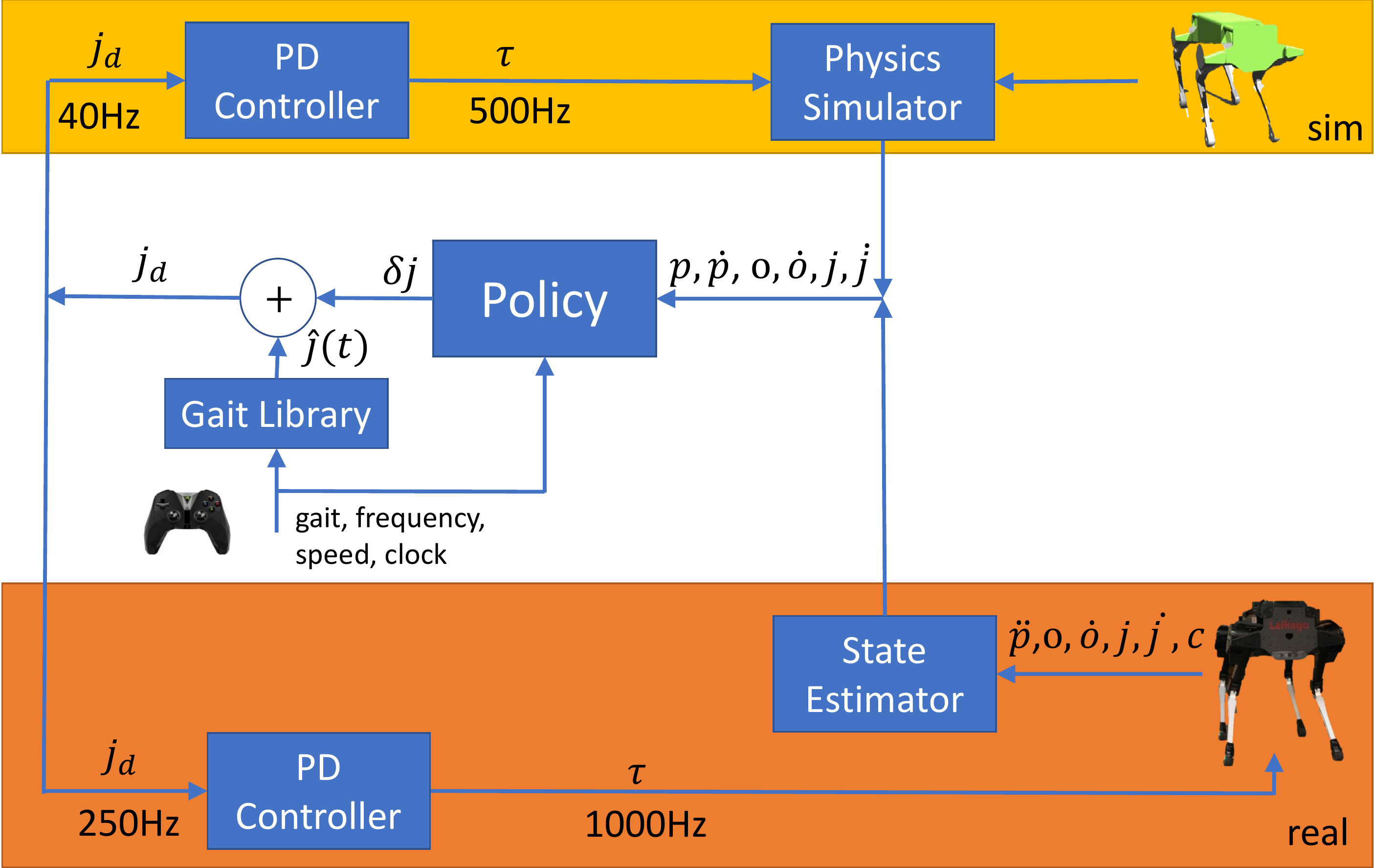}
  \caption{Overview of our system. The input to the policy includes robot state and user commands. The output is a residual PD target, which is added to a reference target and applied to a joint PD controller. Various motions are achieved by using a library of gaits as reference trajectories.}
  \label{fig:system}
  \vspace{-10pt}
\end{figure}

We first describe the structure of our control policies together with the training methods. 
We denote the state of the robot, $x$, as the collection of tuples $x = [p \in \mathbb R^3, o \in S^3, j \in \mathbb R^{12},  \dot{p}, \dot{o}, \dot{j}]$, where $p$ and $o$ are the position and orientation of the base of the robot, and $j$ is the set of twelve joint angles. An overview of our system setup in simulation and on the physical robot is shown in Fig.~\ref{fig:system}. In this section, we describe each component in detail.

\subsection{Training Environment}
We develop policies that can produce gaits that are typical for model-based control of quadrupedal robots. 
We follow an approach similar to prior work~\cite{2020-RSS-laikago} and incorporate reference trajectories into our framework.
Given a reference trajectory $\chi = \{\hat{x}_0, \hat{x}_1, \cdots\}$, where $\hat{x}_t$ is the desired state of the robot at time step $t$, we define the reward $r_t$ as 
   $r_t = 0.4r^j_t + 0.3r^p_t + 0.3r^o_t,$
where 
   $$r^j_t = \exp(-2\norm{j_t-\hat{j}_t}^2),$$
   $$r^p_t=\exp(-\norm{p_t-\hat{p}_t}^2),$$
   $$r^o_t=\exp(-2\norm{o_t-\hat{o}_t}^2-5\norm{\dot{o}_t-\dot{\hat{o}}_t}^2).$$
The goal of RL is to find a policy that will generate action $a_t$ at each time step $t$, such that the cumulative reward $\sum_{t=0}^\infty \gamma^t r_t$ is maximized.

As is common practice for RL with legged robots, the input to the policy includes the current state of the robot but excludes the $x, y$ coordinates, since their measurement tends to drift for the physical robot without exteroceptive sensors\footnote{In practice, the location estimates will of course be significantly dependent on the state estimation scheme, such as the use of leg odometry.}. To achieve a cyclic locomotion gait, we use a reference motion that is a crude cyclic motion sketch  with a period $T$. This consists of sinusoidal motions lifting the feet, and a body moving at constant speed. 
To inform the policy about the time-varying nature of the reward, we also provide inputs $\cos(2\pi t/T)$ and $\sin(2\pi t/T)$ to the policy. 
Finally, we include the gait type (encoded as an integer index), the gait frequency, and the desired speed as inputs.  These are controlled by a human operator to switch between different motions. Following~\cite{2018-IROS-cassie}, the output of the policy is a residual PD joint target $\delta_j$, and the corrected joint position $j_{d} = \delta_j + \hat{j}$ is then used as the input to a PD controller that generates torques $\tau$ on the robot via $\tau = k_p (j_{d}-j) - k_d\dot{j}$. We use $k_p=40$ and $k_d=0.5$ as default gains. 
During training in simulation, the policy is queried at 40\,Hz.
while the PD controller updates the commanded torque at 500\,Hz.

\subsection{Gaits}
We achieve multiple gaits via gait-specific reference trajectories, which provide a sketch of
the desired gait phases.
Different gaits are characterized by their contact sequences. 
The motion is divided into phases, with each phase
having designated legs in either swing or stance roles.  For the stance legs, the reference joint angles are fixed to $[0, 0.65, -1]$ for hip abduction, hip pitch, and knee angles, respectively (in radians).  For the swing legs, the reference angles are $[0, 0.65-0.4 \hat{v}_{x}\sin(\pi \rho/T_p),-1+0.7\sin(\pi \rho/T_p)]$, where $\rho$ is the time elapsed during the current phase, $T_p$ is the duration of the phase, and $\hat{v}_{x}$ is the desired speed in the forward direction that the policy should achieve. The swing trajectories are empirically designed so that the swing foot has high clearance while also going forward or backward to adapt to the desired velocity.

We adopt three common locomotion gaits for quadrupedal robots: 
(1) \textit{Walking}, where each individual leg moves in turn.
(2) \textit{Trotting}, where diagonal legs share a common swing phase;
(3) \textit{Pacing}, where front-and-hind legs on a given side of the body share a common swing phase. 
We further increase the variety of trajectories by varying desired speed $\hat{v}_{x}$ and phase duration $T_p$ to obtain 
a library of gaits that move at different speeds and stepping frequencies.
While we demonstrate successful sim-to-real transfer with all three gaits, 
we focus our evaluation on the \textit{Trotting} and \textit{Pacing}.

\subsection{Training Setup}
We train our policies with actor-critic using proximal policy optimization \cite{2017-arxiv-PPO}. The policy is represented as a two-layer feedforward neural network with a hidden layer size of $128$. During training, the actions follow a Gaussian distribution with mean given by the network output and a fixed standard deviation of $\exp(-2.5)$. During testing, we use the deterministic output, as given by the mean.

We use Isaac Gym during training, which is supported by a GPU-accelerated simulator~\cite{2020-IsaacGym}. This simulator has been validated to simulate rigid body dynamics with reasonable accuracy for locomotion \cite{2020-CoRL-ContactAdapt} and manipulation \cite{2019-icra-phyx2}. We simulate $1600$ robots in parallel and collect $2.4\times 10^5$ environment tuples at each training iteration. We train each policy for a maximum of $10^3$ iterations, with $2.4\times 10^8$ samples in total, less than or comparable to the values used for related sim-to-real work for similar scale quadrupeds, e.g., \cite{2020-RSS-laikago, 2019-science-sim2realAnyaml, 2020-icra-constraintQuadruped}. The training time for one policy on a single GeForce RTX 2080 Ti GPU is $6$ to $8$ hours.

\subsection{Physical Robot Setup}
While the body position and velocity are available for the robot in simulation, we can only estimate these quantities for the physical robot via its onboard sensors, which consist of an IMU, joint encoders, and a one-dimensional force sensor on each foot for contact detection.  We follow prior work~\cite{2018-iros-cheetah3Design} 
and build a Kalman filter for the purpose of state estimation. 

The estimated body velocity can suffer from bias due to integrated accelerometer error. This results in the robot drifting in the plane while being commanded to step in place. This can be addressed by adding an artificial offset to the estimates online to counter the drift. We also add an offset to the yaw angle to compensate for initialization error and drift. These offsets can be also used as a command signal for moving sideways or turning, even though the policy has never explicitly been trained for these motions. For example, if we add a positive offset to the lateral velocity, the robot will move in the negative direction with speed matching the offset in order to make the overall lateral velocity observation zero. These aspects of control arise implicitly from encountering similar states during training due to the stochastic policy.

The target joint angles from the policy are updated at a slow rate, every 26~ms, during training, which improves learning efficiency. In the physical robot experiment, however, the slow update limits the control bandwidth of the PD controller and introduces a discontinuity that can harm the motors due to the large torque change. We mitigate this issue by updating the target joint angles every 4~ms on the physical robot. This is possible since the policy query time typically takes around 2~to 3~ms. To further smooth the target joint angles, we pass them through a discrete first-order low-pass filter before providing them to the PD controller. This produces smoother movements and therefore also helps realize improved state estimates from the robot. The low-pass filter $j_d = (1-\lambda) j_{d, prev} + \lambda j_{d, policy}$ averages the previous target joint angles $j_{d, prev}$ and the current target joint angles $j_{d, policy}$ with a filter constant $\lambda\in[0,1]$. We use a weak filter with $\lambda=0.2$ and a cutoff frequency of 62.5~Hz. A similar smoothing procedure was also used in \cite{2020-RSS-laikago}.

\section{Dynamic Randomization is not Necessary}
The reality gap often causes direct policy transfer to fail. While dynamics randomization is frequently used to address this issue, the randomization is often applied in an {\em ad hoc} fashion, without a deeper
understanding of other possible sources of failure, of the need (or possible lack thereof)
for randomization, or of the impact of randomization choices, including unintended consequences. 
In this section, we demonstrate that dynamics randomization is not necessary for our learned controllers. 
We first show in a simulation that the default policy trained without dynamics randomization can already cope with modeling errors that are larger than the randomization variations used in prior sim-to-real work~\cite{2020-RSS-laikago}. We also then run these policies on the physical robot and demonstrate successful sim-to-real transfer.  

\subsection{Types of Perturbation}
We first describe the types of perturbation we use for testing the robustness of the policies, both in simulation and on the physical robot. A policy is deemed to successfully pass the robustness tests if it can stay balanced for more than 10 seconds. Note that it does not reflect how well the robot does in term of achieving good rewards, e.g., the robot can balance while failing to follow the command speed. We do not document tests for robustness to terrain friction, as in practice we have not found this to be a significant issue for sim-to-real transfer.

\paragraph{Mass Perturbation}
One source of modeling error is given by the mass and moments of inertia of the body parts. These are also the most commonly used parameters for dynamics randomization. In simulation, we directly change the mass of the main body and record the maximum mass value we can add to the default value. We put a box of bricks on top of the physical robot and record the maximum payload the robot can carry without falling.

\paragraph{Proportional Gain}
We use $k_p=40$ across all motors during training. During testing in simulation, we decrease the value of $k_p$ and record the minimum value the policy can cope with. This is a proxy for possible friction in the motors as well as understanding
the sensitivity of the learned policy to PD controller parameters. The same procedure is used for physical robot testing.

\paragraph{Latency}
Another commonly randomized parameter is the latency in the system. We simulate the latency by implementing a first-in-first-out buffer, where the length of the buffer corresponds to the introduced latency. During testing, both in simulation and on the physical robot, we record the maximum latency the policy can cope with. 

\paragraph{Lateral Push}
In the simulation, we apply a constant lateral push for $5$ seconds and record the maximum push the policy can recover from. We only perform qualitative tests on the physical robot since 
we do not currently have the hardware needed to apply a prescribed force or impulse.

\paragraph{Slope}
We command the robot to walk up or down a slope and record the maximum slope the policy can walk on without falling.

\subsection{Robustness Tests in Simulation}
We train policies that can perform trotting and pacing at various speeds without dynamics randomization. We deploy the policies in increasingly challenging scenarios for each of the five varieties of perturbation described in the previous subsection, recording the maximum perturbation (e.g., the largest slope angle) at which the robot can still complete the task.  We repeat the experiment by training a total of three policies for each of the two gaits, computing the mean and sample standard deviation of the threshold values over the three trials.  These results are shown in TABLE~\ref{tab:robustness_test}. The policies can already cope with a larger range of perturbations as compared to ranges typically used in related sim-to-real literature for similar robots~\cite{2020-RSS-laikago, 2019-science-sim2realAnyaml}. For example, the robot can carry mass that is up to twice the body mass for the pace,  while 20 percent randomization is typically used. These ranges also cover the possible modeling error. For example, the default value of the mass of the body is measured by the manufacturer and typically has less than a 5 percent error. The latency in the system is mainly caused by the policy query and is typically less than 4~ms, significantly smaller than what the policy can handle in simulation.

\subsection{Tests on the Physical Robot}
\begin{figure}[t]
  \centering
  \includegraphics[width=\columnwidth]{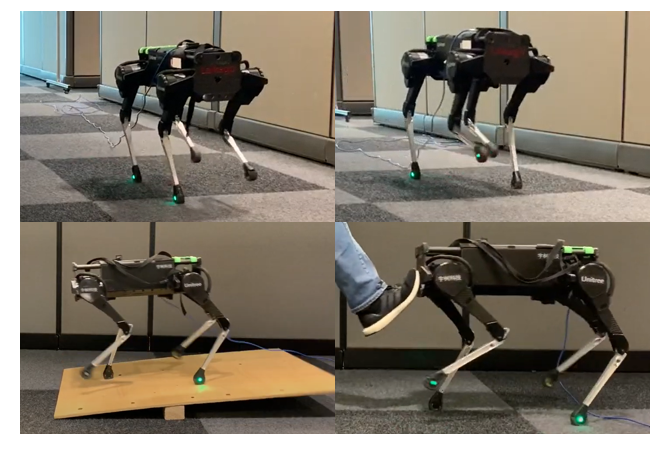}
  \caption{We train policies for multiple gaits that transfer to the real robot without dynamics randomization. \textbf{TOP}: Trotting and pacing with learned policies. \textbf{BOTTOM}: The policies can walk up or down a slope and react to forceful pushes despite never encountering these scenarios during training.}
  \label{fig:sim-to-real}
  \vspace{-15pt}
\end{figure}

We directly apply these policies on the physical robot and observe sim-to-real success without adaptation. Fig.~\ref{fig:sim-to-real} shows snapshots of the physical robot tests. This is consistent with our observation in the simulation where the policies are capable of coping with large modeling errors.

We perform similar robustness tests on the physical robot for the trotting and pacing gaits. 
Due to concern over potential damage to the hardware, we only test one of the policies for each gait. 
The results are recorded in TABLE~\ref{tab:robustness_test}. 
For the trotting gait, the reality gap is smaller as the robustness test results are similar between simulation and the physical robot, while for the pacing gait, we observe that the policy typically achieves worse performance on the physical robot compared to the simulation robustness test. However, the policies are nevertheless already robust against a large range of perturbations, and in both cases, the reality gap is easily overcome. As observed here, the reality gap is also dependent on the type of motions considered and thus the results may vary for more dynamic motions such as gallops or jumps.

\begin{table*}[t]
\begin{center}
\begin{tabular}{|l|c|c|c|c|c|c|}
\hline
\textbf{Policy} & \textbf{$\Delta$Mass (kg)} & \textbf{P gain (Nm/rad)} & \textbf{Latency (ms)} & \textbf{Lateral Push (N)} & \textbf{Slope Up (degrees)} & \textbf{Slope Down (degrees)} \\
\hline
Trot: Simulation &$9\pm3$&$27\pm2$&$17\pm1$& $50\pm7$&$11\pm1$&$6\pm0$ \\
\hline
Trot: Real & $10$&$27$&$16$& not measured &$4$&$6$\\
\hline
Pace: Simulation &$20\pm3$&$23\pm1$&$17\pm1$ &$43\pm2$&$13\pm1$&$11\pm0$\\
\hline
Pace: Real & $8$&$30$&$16$& not measured &$4$&$6$\\
\hline
\end{tabular}
\end{center}
\caption{Robustness tests of different controllers in simulation and in experiments with the real robot.}
\label{tab:robustness_test}
\vspace{-10pt}
\end{table*}

\section{Dynamics Randomization is not Sufficient}
In the previous section we demonstrated that dynamics randomization is, in the right circumstances, not necessary for direct sim-to-real transfer. This contrasts with previous work~\cite{2020-RSS-laikago} despite being similar in the use of reference trajectories, being evaluated on similar classes of motions (trotting and pacing), and on the same robot model. We observe several design decision discrepancies and conduct an ablation study on these design decisions. We thus provide an explanation for the apparent contradiction and show that with alternate design choices, control policies will fail to transfer and that dynamics randomization by itself, without further on-robot adaptation, is not sufficient to cross the reality gap in these scenarios. This is consistent with other sim-to-real results, where an additional adaptation step on the physical robot is needed \cite{2019-iros-sim2realbiped, 2020-RSS-laikago}.

\subsection{Design Choices}
\paragraph{Choice of Observation}
In \cite{2020-RSS-laikago}, the state observations consists of raw sensory measurements such as the orientation of the body and joint angles, while we use a state estimator and include additional information such as body velocity and joint velocity. To understand the impact of this design choice, we train policies without the body velocity, to create a more closely matched system.
While some velocity information is implicit in the proprioceptive state, we hypothesize that a good estimate of the body velocity can help prevent drift and instability in the lateral direction.

\paragraph{Choice of Proportional Gain}
In \cite{2020-RSS-laikago}, the authors use stiff proportional gains ($k_p=220$) for the PD controllers on the motors, while we use a soft gain ($k_p=40$). We conduct experiments to evaluate the impact of this design choice.

\subsection{Simulation Test}
We train four different policies to explore the impact of the described design choices.
First, we train pacing policies without velocity feedback, with and without dynamics randomization.
Second, we train trotting policies with a high gain ($k_p=160$), with and without dynamics randomization. 
The type and range of randomization can be seen in Table~\ref{tab:randomization_param}. 

For the high-PD-gain trot, we use $k_p=160$ instead of $k_p=220$ as used in \cite{2020-RSS-laikago}
because the higher gain can cause instability on the robot.
We also purposely choose not to apply randomization to lateral pushes or slopes during training.
Instead, we wish to understand cross-correlations in robustness.
Specifically, if we train for additional robustness along some dimensions or parameters,
via dynamics randomization, will that positively or negatively influence the
robustness along the other unrandomized dimensions? 

We train three different policies under each setting and perform robustness tests similar to the previous section; the results are shown in TABLE~\ref{tab:design_comparison}. We observe that dynamics randomization does not help the observed robustness in most of the robustness tests, including in the randomized dimensions.
Thus, rather surprisingly, randomization does not generally help robustness in our experiments.
The one exception is latency randomization, which was also significantly randomized in \cite{2020-RSS-laikago}. Overall, this points to much randomization being unnecessary or even harmful.

\begin{table}[t]
\begin{center}
\begin{tabular}{|l|c|}
\hline
\textbf{Parameter} & \textbf{Range} \\
\hline
Mass & $[0.8, 1.2]\times \text{default}$\\
\hline
Inertia & $[0.5, 1.5]\times \text{default}$\\
\hline
P gain & $[-20\,\text{Nm/rad}, 20\,\text{Nm/rad}] + \text{default}$\\
\hline
Latency & $[0\,\text{ms}, 20\,\text{ms}]$ \\
\hline
\end{tabular}
\end{center}
\caption{Randomized parameters and their ranges.}
\label{tab:randomization_param}
\vspace{-15pt}
\end{table}

\subsection{Robot Test}
We also verify that the policies fail to transfer to the physical robot with the alternative design choices, even with dynamics randomization. This is consistent with prior results~\cite{2020-RSS-laikago}, where additional adaptation procedures were employed on the physical robot to cross the reality gap.

The pacing policies without velocity feedback cannot control the lateral velocity and often fall sideways. 
In an additional experiment, we train pacing policies with lateral pushes applied during training.  We find, however, that this cannot compensate for the lack of lateral velocity feedback, and the policies fail in a similar way.

The trotting policies trained with $k_p=160$ exhibit very stiff motion compared to those trained with $k_p=40$, and the motion is sensitive to contact timing. The robot occasionally kicks its feet backward when the policy expects them to be in the stance phase while the foot has yet to make contact with the ground.
This is typically a disadvantage of position control where unexpected contact modes can be problematic. 

\subsection{Comparison of Proportional Gain}
\begin{figure}[t]
      \centering
    \includegraphics[width=0.95\columnwidth]{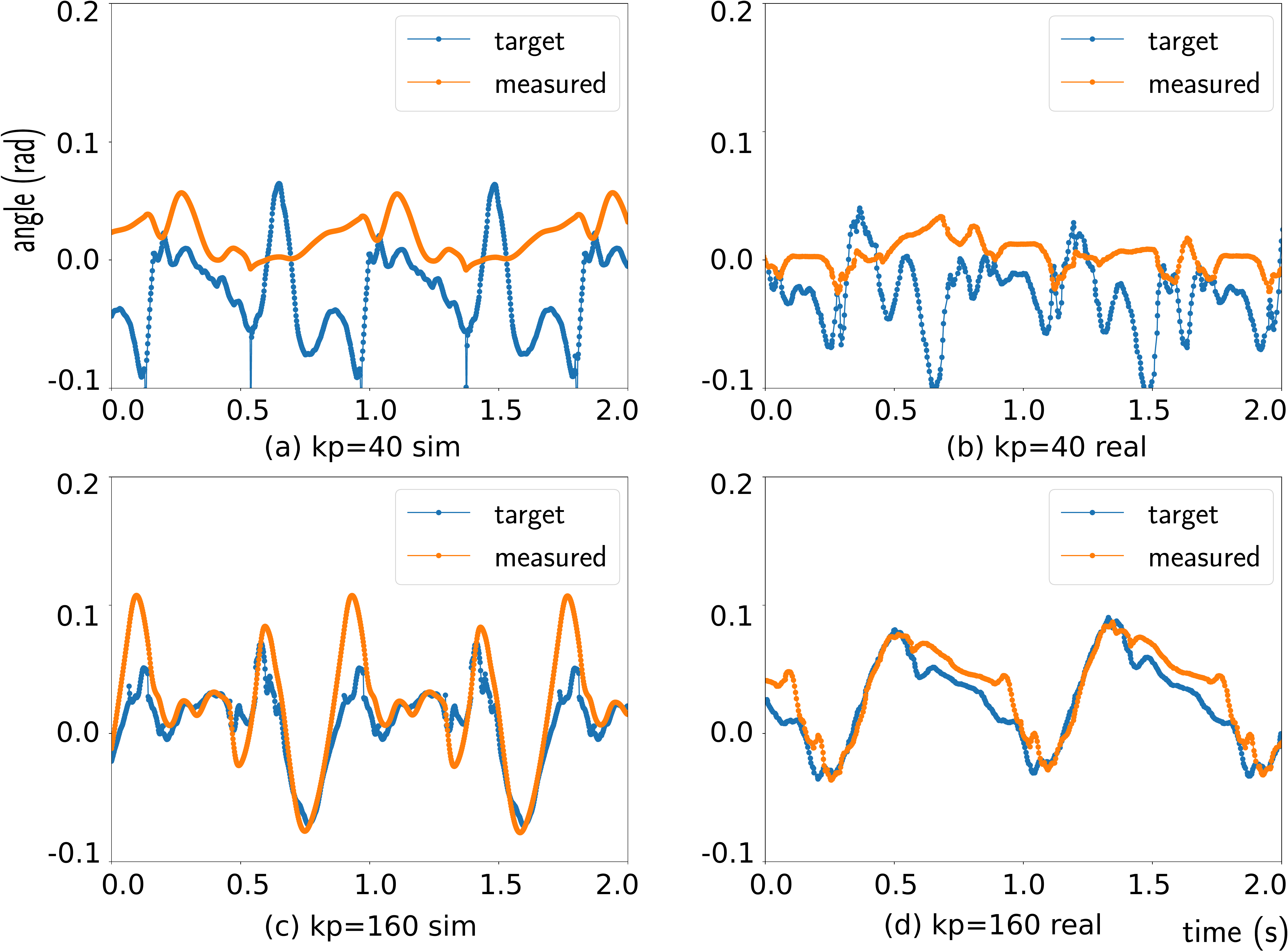}
      \caption{Comparison of different $k_p$. \textbf{TOP}: With $k_p=40$, the tracking error is large. The policy behaves like a torque controller with relatively small reality gap. \textbf{BOTTOM}: With $k_p=160$, the tracking error is much smaller. The policy behaves like a position controller with larger reality gap.}
      \label{fig:P_comparison}
      \vspace{-15pt}
\end{figure}

Inspired by a comment in~\cite{2019-science-sim2realAnyaml}, we hypothesize that low proportional gains will result in a policy behaving like a torque controller, i.e., using the target angle as a proxy to achieve a desired force, while high proportional gains will behave like a position controller. Related work shows that a torque controller is typically more robust than a position controller for unanticipated impacts~\cite{2012-AMC-hyqImpedance}. In Fig.~\ref{fig:P_comparison}, we plot the joint and joint-target trajectories for a hip joint and observe that with $k_p=40$, the policy generates (undisturbed) motions with large PD-tracking errors and thus behaves more like a torque controller, while with $k_p=160$, the policy generates motions with low tracking errors and behaves like a position controller. 

\definecolor{darkgreen}{RGB}{0,160,0}

\begin{table*}[t]
\begin{center}
\begin{tabular}{|l|c|c|c|c|c|c|c|}
\hline
 & \textbf{Mass} & \textbf{P gain} & \textbf{Latency} & \textbf{Lateral Push} & \textbf{Slope Up} & \textbf{Slope Down} & \textbf{Sim-to-Real} \\
\textbf{Policy} & \textbf{(kg)} & \textbf{(Nm/rad)} & \textbf{(ms)} & \textbf{(N)} & \textbf{(degrees)} & \textbf{(degrees)} & \textbf{Outcome} \\
\hline
Pace: Default&$20\pm3$& $23\pm1$&$17\pm1$ &$43\pm2$&$13\pm1$&$11\pm0$&success\\
\hline
Pace: No vel &\color{blue}$18\pm0$ &\color{blue}$24\pm2$ &\color{red}$12\pm2$&\color{red}$22\pm0$ &\color{red}$4\pm0$&\color{red}$7\pm0$&\color{red}failure\\
\hline
Pace: No vel, with rand & \color{red}$9\pm4$&\color{red}$30\pm2$&\color{darkgreen}$38\pm2$&\color{red}$13\pm5$&\color{red}$4\pm1$&\color{red}$9\pm1$&\color{red}failure\\
\hline
Trot: Default & $9\pm3$& $27\pm2$ & $17\pm1$ & $50\pm7$ & $11\pm1$ & $6\pm0$ & success\\
\hline
Trot: $k_p=160$ &\color{blue}$18\pm12$&$--$&\color{blue}$17\pm4$ &\color{red}$18\pm6$&\color{blue}$10\pm0$&\color{blue}$5\pm3$&\color{red}failure \\
\hline
Trot: $k_p=160$, with rand &\color{blue}$8\pm1$&$--$&\color{darkgreen}$41\pm1$&\color{red}$12\pm8$&\color{blue}$12\pm1$&\color{red}$1\pm0$&\color{red}failure\\
\hline
\end{tabular}
\end{center}
\caption{Robustness tests for policies trained under different setup, together with the result of attempted sim-to-real transfer. \textcolor{blue}{Blue}
indicates policies that perform similarly to the corresponding policy with default settings. \textcolor{darkgreen}{Green} and \textcolor{red}{red} indicate policies that perform better or worse than the default, respectively.  Policies without velocity feedback or with $k_p=160$ all fail the sim-to-real tests. They also generally perform worse in the robustness tests compared to default.}
\label{tab:design_comparison}
\vspace{-15pt}
\end{table*}

\section{Effect of Dynamics Randomization}
We have observed that dynamics randomization is neither necessary in our setting, 
nor sufficient in the face of other problematic design choices.
However, different conclusions might be drawn for different robots or different motions.
In this section, we explore the advantages and disadvantages of dynamics randomization in greater depth.


\subsection{Dynamics Randomization Produces Conservative Policies}
We observe in TABLE~\ref{tab:design_comparison} that dynamics randomization can sometimes lead to policies that are overly-conservative in order to achieve unnecessary robustness in parameters that are being randomized. For example, the pacing policies trained with no velocity feedback and dynamics randomization perform worse than policies trained without dynamics randomization in general, except in terms of dealing with latency. However, the physical robot system has an estimated latency of less than 4~ms, and this unnecessary robustness against increased latency leads to compromised performance and robustness along other dimensions.

We further train trotting policies under the default setting with dynamics randomization. We observe a more conservative maximum speed (0.9\,m/s with randomization and 1.1\,m/s with no randomization), both in simulation and on the physical robot. This also corresponds to our intuition that dynamics randomization can produce conservative policies.

\subsection{Randomize Parameters that Matter}
We use the latency test to investigate the usefulness of dynamics randomization. We observe that policies trained without randomization fail when the latency exceeds 17~ms. We train another policy with randomized latency only; more specifically, the policy is trained with randomized latency of up to 20~ms. The resulting policy can handle latency up to 32~ms, 
both in simulation and on the physical robot. 

This indicates that dynamics randomization can help in scenarios where significant modeling errors are present, such as latency in the system. In these scenarios, dynamics randomization provides a useful mechanism to cross the reality gap
by only randomizing the parameters that are responsible. 

\subsection{Summary}
We observe that blindly applying dynamics randomization when it is not necessary can generate suboptimal policies that are too conservative. However, if the system has fundamental modeling errors that hinder sim-to-real success, randomization is needed to cross the reality gap, as shown in our latency experiments. We note that actuator modeling errors can also pose a sim-to-real challenge, as noted in \cite{2019-science-sim2realAnyaml}, where a learned actuator model is employed to cross the reality gap. 

In summary, we suggest employing dynamic randomization or additional modeling only when significant modeling errors are present and to only randomize or model parameters that matter. Superfluous dynamics randomization harms performance in measurable ways while possibly giving no extra benefit in robustness, even for the randomized dimensions.

\section{Discussion and Future Work}

In this paper, we have studied a number of factors that affect sim-to-real transfer for quadrupedal locomotion, and found that commonly-used dynamics randomization often offers negligible actual improvements in robustness.  We have also evaluated design choices related to proportional gain stiffness and state observation parameters that may have been overlooked in prior work, despite their critical role in the success of sim-to-real transfer.

While many of these observations should generalize beyond our particular settings, additional modeling or randomization may be necessary for other tasks or robot morphologies. For example, due to different actuation, the ANYmal robot has been found to require additional actuator modeling~\cite{2019-science-sim2realAnyaml}, and significant latency in the control loop of the Ghost Minitaur robot has been found to require randomization of latency~\cite{2018-rss-sim2realquadruped}. We advocate identifying important, i.e., high-sensitivity, sim-to-real bottlenecks using simulations and performing necessary additional modeling or randomization only for the relevant parameters instead of arbitrarily adding randomization to a larger set of parameters, as has sometimes been done in the past.

It is likely that the reality gap will be more pronounced for more dynamic motions such as running or jumping. 
We plan to improve our understanding of sim-to-real challenges that might be present in such settings.
Relatedly, we wish to identify the key design choices for sim-to-real success in more general settings.  In this work, we tuned certain parameters (such as proportional gains) empirically, while recognizing that such design considerations can vary from robot to robot.  

Recent work also utilizes state history~\cite{2020-science-blindQuadruped} or recurrent neural network~\cite{2020-rss-RNNCassie} so that the policy can cope with challenging terrains without using additional exteroceptive sensing, which cannot be achieved with our current setup. Randomization during training in these cases are also necessary to train an adaptive policy. However, we note that the randomization is applied to the external environment rather than the robot parameters. This thus remains in accordance with our premise that one should only randomize a minimal set of parameters.






\bibliographystyle{IEEEtran}
\bibliography{icra}

\end{document}